\documentclass{llncs}
\usepackage{llncsdoc,amsmath,graphicx}
\usepackage{algorithm}
\usepackage{mathtools}
\usepackage{amssymb}
\usepackage{amsmath}
\usepackage{graphicx}
\usepackage{subcaption}
\usepackage{algorithm}
\usepackage[noend]{algpseudocode}
\usepackage[misc]{ifsym}

\def\x{{\mathbf x}}

\title{A Discriminative Event Based Model for Alzheimer's Disease Progression Modeling}

\begin{document}
\date{}
\author{}
\institute{}
\author{Vikram Venkatraghavan \inst{1} \textsuperscript{\Letter} \and
Esther  E. Bron \inst{1} \and
Wiro J. Niessen \inst{1,2} \and
\mbox{Stefan Klein} \inst{1} }
\institute{$^1$ Biomedical Imaging Group Rotterdam, Depts. of Medical Informatics \& Radiology, Erasmus MC, The Netherlands \\ 
$^2$ Faculty of Applied Sciences, Delft University of Technology, The Netherlands \\
\email{v.venkatraghavan@erasmusmc.nl} }

\maketitle

\begin{abstract}

The event-based model (EBM) for data-driven disease progression modeling estimates the sequence in which biomarkers for a disease become abnormal. This helps in understanding the dynamics of disease progression and facilitates early diagnosis by staging patients on a disease progression timeline. Existing EBM methods are all generative in nature. In this work we propose a novel discriminative approach to EBM, which is shown to be more accurate as well as computationally more efficient than existing state-of-the art EBM methods. The method first estimates for each subject an approximate ordering of events, by ranking the posterior probabilities of individual biomarkers being abnormal. Subsequently, the central ordering over all subjects is estimated by fitting a generalized Mallows model to these approximate subject-specific orderings based on a novel probabilistic Kendall's Tau distance. To evaluate the accuracy, we performed extensive experiments on synthetic data simulating the progression of Alzheimer's disease. Subsequently, the method was applied to the Alzheimer's Disease Neuroimaging Initiative (ADNI) data to estimate the central event ordering in the dataset. The experiments benchmark the accuracy of the new model under various conditions and compare it with existing state-of-the-art EBM methods. The results indicate that discriminative EBM could be a simple and elegant approach to disease progression modeling.

\end{abstract}
%
%
\section{Introduction} \label{sec:intro}

Alzheimer's Disease (AD) is characterized by a cascade of biomarkers becoming abnormal, the pathophysiology of which is very complex and largely unknown. However, understanding the progression of several imaging and clinical biomarkers after disease onset is extremely important for both early diagnosis and patient staging.
Conventional models of disease progression reconstruct biomarker trajectories in individual subjects using longitudinal data. This is done to get insight into disease progression mechanism~\cite{Sabuncu:2014}~\cite{Schmidt:2016}. However, the utility of such models are restricted by the fact that longitudinal data in large groups of patients is scarce. To circumvent this problem, methods to infer the order in which biomarkers for a disease become abnormal based on cross-sectional data have been proposed~\cite{Fonteijn:2012}~\cite{Medina:2016}. Some of these models~\cite{Jack:2013}~\cite{Medina:2016} rely on stratification of patients into several subgroups based on symptomatic staging, for inferring the aforementioned ordering. However, the problem with using symptomatic staging is that it is very coarse and qualitative.

Event based modeling (EBM)~\cite{Fonteijn:2012}~\cite{Huang:2012}~\cite{Young:2014} is a data-driven approach to disease progression modeling. EBM algorithms neither rely on symptomatic staging nor on the presence of longitudinal data for inferring the temporal ordering of events, where an event is defined by a biomarker becoming abnormal. All the variants of EBM developed so far are generative in nature. The existing state-of-the-art EBM methods are either not very robust in handling disease heterogeneity or scalable to large number of biomarkers, as will be demonstrated in this paper.

In this work, we propose a novel discriminative approach to EBM (DEBM) to address these issues. We assume that the event orderings for each subject in the dataset need not be unique, but form a cluster around a single central ordering. We first compute a noisy estimate of event ordering for each subject by ranking the posterior probabilities of individual biomarkers being abnormal. Subsequently, we introduce a novel probabilistic Kendall's Tau distance to reliably aggregate such noisy subject-specific event orderings to estimate a central ordering over all subjects.


\section{Event Based Models} \label{sec:relatedwork}

The EBM considers disease progression as a series of events,
where each event corresponds to a new biomarker becoming abnormal. Fonteijn's EBM~\cite{Fonteijn:2012} finds the ordering of events $(\sigma_0)$ such that the likelihood that a dataset was generated from subjects following this event ordering is maximized. This event ordering $(\sigma_0)$ consists of a discrete set of events $\{E_{\sigma_0(1)}, E_{\sigma_0(2)}, ..., E_{\sigma_0(N)}\}$, where $N$ is the number of biomarkers per subject in the dataset. 

Given a cross-sectional dataset of $M$ subjects, if $X_j$ denotes a measurement of biomarkers for each subject at a certain timepoint, with each measurement $X_j$ consisting of $N$ biomarker values $\x_{j,i}$, the likelihood of the dataset being generated by $\sigma_0$ as defined by~\cite{Fonteijn:2012} is given by:

\begin{equation}
\label{eq:font1}
p\left( X|\sigma_0 \right) = \prod_{j=1}^{M} \sum_{k=0}^N p(k) \left( \prod_{i=1}^k p \left( x_{j,\sigma_0(i)} | E_{\sigma_0(i)} \right) \prod_{i=k+1}^N p \left( x_{j,\sigma_0(i)} | \neg E_{\sigma_0(i)} \right) \right)
\end{equation}

\noindent where $p(k)$ is the prior probability of a subject being at position $k$ of the event ordering, which is assumed to be equal for each position. With the assumption that all the biomarkers in the control population are normal and that the biomarker values follow a Gaussian distribution, $p \left( x_{j,\sigma_0(i)} | \neg E_{\sigma_0(i)} \right)$ is computed. Abnormal biomarker values in the patient population are assumed to follow a uniform distribution but not all biomarkers of a patient could be assumed to be abnormal. For this reason, the likelihoods were obtained using a mixture model of Gaussian-uniform distributions where only the parameters of the uniform distribution were allowed to be optimized.

This method was slightly modified in~\cite{Young:2014} to estimate the optimal ordering in a sporadic AD dataset with significant proportions of controls were expected to have presymptomatic AD. A Gaussian distribution was used to describe both the control and patient population, and the mixture model allowed for optimization of parameters for the Gaussians describing both control and patient population. Gaussian mixture model was also used to incorporate more subjects from the dataset with clinical diagnosis of mild cognitive impairment (MCI).

An assumption made in~\cite{Fonteijn:2012} and~\cite{Young:2014} is the existence of a single ordering common in all the subjects within a cohort. Such as assumption is rather too restrictive for estimating the progression of a complex disease such as AD. This assumption was relaxed in~\cite{Huang:2012}, which estimates a distribution of event orderings with a central event ordering and a spread as per a generalized mallows model~\cite{Fligner:1988}. Huang's EBM~\cite{Huang:2012} is an expectation maximization algorithm to obtain the central ordering $\sigma_0$ and spread $\phi$. The E-step estimates the likelihood of a patient's biomarker value measurement following an event order $\sigma_j$, given $\sigma_0$ and $\phi$. In the M-step, $\sigma_0$ and $\phi$ are estimated based on $(\sigma_j)$ estimated in the E-step. This is done iteratively to maximize the likelihood of generation of patients' data based on $\sigma_0$ and $\phi$.

While Fonteijn's EBM is computationally inexpensive, the assumptions are very restrictive. The assumptions in Huang's EBM on the other hand are realistic, however the algorithm does not scale well to large number of biomarkers. With these in mind, we propose a discriminative approach to EBM.

\section{Discriminative Event Based Model} \label{sec:DEBM}

In this section, we propose our novel method for estimating central ordering of events $(\sigma_0)$. We postulate that the posterior probability of a biomarker being abnormal signifies the progression of a biomarker. Since this is done for each biomarker measured from a subject, the different amounts of progression for different biomarkers as estimated by their corresponding posterior probabilities signify the event ordering in a subject $(\sigma_j)$. However, the posterior probability is not only affected by progression of the biomarker to its abnormal state, but also by inherent variability in `healthy' biomarker values across subjects, and by measurement noise. Since it is not feasible to distinguish between these effects based on a single (cross-sectional) measurement, we expect $\sigma_j$ to be a noisy estimate. To estimate $\sigma_0$ based on noisy estimates of $\sigma_j$, we introduce a novel variant of Kendall's Tau distance which takes into account the posterior probability estimates. The proposed framework is discriminative in nature, since we estimate $\sigma_j$ directly based on the posterior probabilities of individual biomarker becoming abnormal. This is in contrast to the existing EBM models, which estimates the event orderings based on the likelihood of the data being generated by an ordering. 

The rest of the section is organized as follows: Given a single cross-sectional measurement of biomarkers from a subject, we present a method to estimate $\sigma_j$ in Section~\ref{ssec:subjectordering}. The problem of estimating $\sigma_0$, from noisy estimates of $\sigma_j$ is addressed in Section~\ref{ssec:CentralOrdering}.

\subsection{Biomarker Progression and Subject-wise Event Ordering} \label{ssec:subjectordering}

Assuming a paradigm similar to that in previous EBM variants~\cite{Huang:2012}~\cite{Young:2014}, the probability density functions (PDF) of normal and abnormal classes in the biomarkers are assumed to be represented by Gaussians. There are two reasons why constructing these PDFs is non-trivial. Firstly, the labels (clinical diagnoses) for the subjects do not necessarily represent the true labels of all the biomarkers extracted from the subject. Not all biomarkers are abnormal for a subject with AD, while some of the controls could have undiagnosed pre-symptomatic conditions. Secondly, the clinical diagnosis is sometimes non-binary and includes classes such as MCI, with significant number of biomarkers in normal and abnormal classes.

In our approach we address these two issues independently. We make an initial estimate of the PDFs using biomarkers from `easy' controls and `easy' AD subjects and later refine the estimated PDF using the entire dataset.

A Bayesian classifier is trained for each biomarker using controls and AD subjects, based on the assumption that there are no wrongly-labeled biomarker in either class. This classifier is subsequently applied to the training data, and the predicted labels are compared with the clinical labels. The misclassified data in the training dataset could either be outliers in each class resulting from using untrustworthy labels or could genuinely belong to their respective classes and represent the tails of the true PDFs. Irrespective of the reason of misclassification, we remove them for initial estimation of the PDFs. This procedure thus results, for each biomarker, in a set of `easy' controls (whose biomarker values represent normal values) and `easy' AD subjects (whose biomarker values represent abnormal values).

As we use Gaussians to represent the PDFs, we have initial estimates for mean and standard deviation for both normal and abnormal classes for each biomarker. We refine these estimates using a Gaussian mixture model (GMM) and include all the available data, including MCI subjects and previously misclassified cases. The objective function for optimization for biomarker $x_{:,i}$ is:

\begin{equation}
\label{eq:GMMopt}
 C_i = \sum_j \log \left[ \left( \theta_i \times p(x_{j,i} | E_{j,i}) \right) + \left( \left(1-\theta_i \right) \times p(x_{j,i} | \neg E_{j,i}) \right) \right]
\end{equation}

\noindent Where $\theta_i$ is the mixing parameter which determines the proportion of abnormal biomarker data in the dataset and lies between $\left[0,1\right]$. To obtain a robust GMM fit, a constrained optimization method is used, putting bounds on the mean and standard deviation parameters. These bounds are set to the $95\%$ confidence interval limits of the initial estimates of means and standard deviation.

The PDF thus obtained is used for classification of the biomarkers using a Bayesian classifier where the mixing parameter $\theta_i$ is used as the prior probability when estimating posterior probabilities for each biomarker. We assume these posterior probabilities to be a measure of progression of a biomarker. Thus, sorting these biomarkers based on decreasing estimates of posterior probabilities results in a noisy estimate for $\sigma_j$. 

\subsection{Estimating the Central Ordering} \label{ssec:CentralOrdering}

Since the event orderings for each subject are estimated independent of each other, any heterogeneity in disease progression is captured in these estimates of $\sigma_j$. The central event ordering $(\sigma_0)$ is the mean of the subject-specific estimates of $\sigma_j$. To describe the distribution of $\sigma_j$, we make use of a generalized Mallows model. The generalized Mallows model is parameterized by a central (`mean') ordering as well as spread parameters (analogous to the standard deviation in a normal distribution). The central ordering is defined as the ordering that minimizes the sum of distances to all subject-wise orderings $\sigma_j$. To measure distance between orderings, an often used measure is Kendall's Tau distance \cite{Huang:2012}. Kendall's Tau distance between a subject specific event ordering $(\sigma_j)$ and central ordering $(\sigma_0)$ can be defined as:

\begin{equation}
\label{eq:kendall21}
K(\sigma_0,\sigma_j) = \sum_{i=1}^{N-1} V_i(\sigma_0,\sigma_j)
\end{equation}

\noindent where $V_i(\sigma_0,\sigma_j)$ is the number of adjacent swaps needed so that event at position $i$ is the same in $\sigma_j$ and $\sigma_0$. 

Since the estimates of $\sigma_j$ are based on rankings of posterior probabilities, it would be desirable to penalize certain swaps more than others, based on how close the posterior probabilities were to each other. To this end, we introduce a probabilistic Kendall's Tau distance, which penalizes each swap based on the difference in posterior probabilities of the corresponding events. The probabilistic Kendall's Tau is computed sequentially using the following algorithm:

\begin{algorithm}[H]
\caption{Probabilistic Kendall Tau distance between Event Orderings}\label{alg:kendtau}
\begin{algorithmic}[1]
\For{$i \in \{1,N-1\}$}
	\State $k \leftarrow \sigma^{-1}_j\left(\sigma_0(i)\right)$
	\If{$ k > i$}
		\State $V_i(\sigma_0,\sigma_j) \leftarrow p \left(E_{j,\sigma_0(k)} | x_{j,\sigma_0(k)} \right) - p \left(E_{j,\sigma_0(i)} | x_{j,\sigma_0(i)} \right)$
		\State Move $\sigma_j(k)$ to position $i$ and update $\sigma_j$
	\Else
		\State $V_i(\sigma_0,\sigma_j) \leftarrow 0$
	\EndIf
\EndFor
\end{algorithmic}
\end{algorithm}

This variant of Kendall's Tau distance is quite close to the weighted Kendall's Tau distance defined in the permutation space introduced in~\cite{Kumar:2010}. The differene stems from the fact that since the probabilistic Kendall's Tau distance is between individual estimates and a central-ordering, the penalization of each swap is weighted assymetrically as $V_i(\sigma_0,\sigma_j) \neq V_i(\sigma_j,\sigma_0)$. This asymetrical weighing can be formulated as a special case of the aforementioned weighted Kendall's Tau distance. 

Computing a global optimum for the central ordering based on subject-wise orderings is NP-hard. The optimization algorithm used in our implementation is based on algorithm introduced by Fligner \textit{et. al.}~\cite{Fligner:1988} to make an unbiased estimate of the central ordering.

\section{Experiments}  \label{sec:exp}

This section describes the experiments performed to benchmark the accuracy of the proposed DEBM algorithm and compare it with state-of-the-art EBM methods. We begin with the details of the experiments performed on ADNI data to estimate the event ordering in Section~\ref{ssec:adniexp}. Such an event ordering serves as a timeline for disease progression and is used for patient staging. Since the groundtruth event ordering is unknown for clinical datasets, we resort to using accuracy of patient staging as an indirect way of measuring the reliability of the event ordering. We also measure the accuracy of event ordering in a much more direct way by performing extensive experiments on synthetic data simulating the progression of AD. The details of these experiments are given in Section~\ref{ssec:simulexp}.

\subsection{ADNI Data} \label{ssec:adniexp}

We considered 509 ADNI\footnote{http://adni.loni.usc.edu/} subjects (162 healthy controls, 210 MCI and 137 AD subjects) who had a 1.5T structural MRI (T1) scan at baseline. The T1w scans were non-uniformity corrected using the N3 algorithm~\cite{Tustison:2010}. This was followed by multi-atlas brain extraction using the method described in Bron \textit{et al.}~\cite{Bron:2014}. Multi-atlas segmentation was performed using the structural MRI scans to obtain a region-labeling for $83$ brain regions in each subject using a set of $30$ atlases~\cite{Hammers:2003}. We calculated the volume of these regions and used the ratio of these volumes with the intra-cranial volume as biomarkers. This is done to compensate for the inter-subject variability in head size. We also downloaded CSF (A$\beta_{1-42}$, tau and p-tau) and cognitive scores (MMSE, ADAS-Cog, RAVLT) biomarker values from the ADNI database. Out of these, volume based biomarkers of $41$ regions, $3$ CSF and $3$ cognitive scores were found to be significant features based on Student's t-test with $p<0.01$. These biomarker values were used to perform three sets of experiments.

\textbf{Experiment 1:} A subset of $7$ biomarkers including the $3$ CSF features, MMSE scores, ADAS-Cog scores, volume of the hippocampus and whole brain was created. Event ordering of these $7$ biomarkers was inferred using DEBM, Huang's EBM~\cite{Huang:2012} and the variant of Fonteijn's EBM that is suited for AD disease progression modeling~\cite{Young:2014}. The original Fonteijn's EBM~\cite{Fonteijn:2012} differs from the version in~\cite{Young:2014} only in the way in which normal and abnormal biomarker distributions are estimated. As an indirect way of measuring the reliability of the estimated event ordering, we use patient staging based on the estimated event orderings as a way to classify controls and AD subjects in the database. $10$-fold cross validation was used for this purpose. AUC measures were used to measure the performance of these classifications and thus indirectly hint at the reliability of the event ordering based on which the corresponding patient staging were performed. We used the patient staging algorithm described in~\cite{Fonteijn:2012} for all three EBMs to ensure that the difference in obtained AUC is strictly because of the obtained event ordering.

\textbf{Experiment 2:} The above experiment was repeated for the entire set of $47$ features. This was done to study the scalability of EBM techniques.

\textbf{Experiment 3:} We studied the positional variance of central ordering inferred by DEBM by creating $100$ bootstrapped samples of the data with $7$ biomarkers followed by computing the central ordering for each of those samples.

\subsection{Simulation Data} \label{ssec:simulexp}

We use the framework developed by Young \textit{et al.}~\cite{Young:2015} for simulating cross-sectional data consisting of scalar biomarker values for healthy controls, MCI and AD subjects. In this framework, disease progression in a subject is indicated by a cascade of biomarkers becoming abnormal and individual biomarker trajectories are represented by a sigmoid. The equation for generating biomarker values for different subjects is given below:

\begin{equation}
\label{eq:sigmoid}
x_{j,i}(\Psi) = \frac{1}{1+ \exp(-\rho_{j,i}(\Psi - \xi_{j,i}))} + \beta_{j,i}
\end{equation}

\noindent $\rho_{j,i}$ signifies the rate of progression of a biomarker with disease state $\Psi$. $\xi_{j,i}$ denotes the disease state at which a biomarker becomes abnormal. $\beta_{j,i}$ denotes the value of the biomarker when the subject is normal. We assume $\rho_{j,i}$ to be equal for all the subjects, for all the biomarkers. With this assumption, variability in a population while simulating a cross-sectional dataset could arise because of variation in either $\beta$ or $\xi$. Variation in $\xi_{j,i}$ results in variation in ordering. In our experiments, $\beta$ and $\xi$ are assumed to be Normal random variables $\mathbb{N}_\beta$ and $\mathbb{N}_\xi$ respectively. Mean of $\mathbb{N}_\beta$ is equal to the mean value of the corresponding biomarker in the controls of the selected ADNI data. We vary relative standard deviation of $\mathbb{N}_\beta$ $(\Sigma_\beta)$ in our experiments, where $1$ refers to the expected variation among healthy controls, estimated based on the selected subjects in ADNI data. Mean of $\mathbb{N}_\xi$ for various biomarkers were assumed to be equi-spaced on the $\Psi$ scale. Standard deviation of $\mathbb{N}_\xi$ $(\Sigma_\xi)$ is varied in multiples of $\Delta \xi / N$, where $\Delta \xi$ is the difference between mean of $\mathbb{N}_\xi$ between adjacent biomarkers.

Using this simulation framework, we study the effect of these two factors in the ability of different variants of EBM algorithms to accurately infer the ground-truth central ordering in the population. Inaccuracy is computed based on the normalized Kendall's Tau distance between the ground truth ordering and most-likely ordering (for Fonteijn's EBM) or central ordering (for DEBM and Huang's EBM). As the Kendall's Tau distance penalizes pair-wise disagreements between event agreements, a normalization factor for $\binom{N}{2}$, where $N$ is the number of events, was chosen to make the accuracy measure interpretable for different number of biomarkers. 

We performed three sets of experiments on simulated data to study the accuracy of the different variants of EBM techniques and several aspects associated with it. For all the experiments, the number of simulated subjects was taken to be equal to the number of subjects in the selected ADNI data. 

\textbf{Experiment 4:} The first experiment was based on selecting a subset of $7$ biomarkers from the $47$ significant biomarkers and study the effect of variation of $\beta$ and $\xi$. For each simulation setting, $50$ repetitions of simulation data were created and used for benchmarking the accuracies of DEBM, Huang's EBM and Fonteijn's EBM.

\textbf{Experiment 5:} The above experiment was repeated for the entire set of $47$ features. This was done to study the scalability of EBM techniques.

\textbf{Experiment 6:} The first experiment was repeated for DEBM and Fonteijn's EBM. In addition to these methods, accuracy of the method using DEBM with normal Kendall's Tau distance and Fonteijn's EBM with the normal and abnormal biomarker distributions estimated based on the method discussed in Section~\ref{ssec:subjectordering} were computed. This was done to ascertain the contributions of individual novel aspects of the proposed algorithm.

\section{Results and Discussions} \label{sec:results}


\subsection{ADNI Data} \label{ssec:adnires}

\begin{figure*}[t!]
    \centering
    \begin{subfigure}[b]{0.5\textwidth}
        \centering
        \label{fig:adni1}
        \includegraphics[width=4cm]{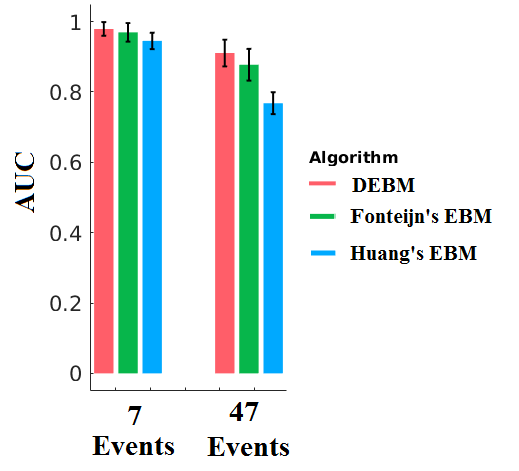}
        \caption{}
    \end{subfigure}%
    ~ 
    \begin{subfigure}[b]{0.5\textwidth}
        \centering
        \label{fig:adni2}
        \includegraphics[width=4.5cm]{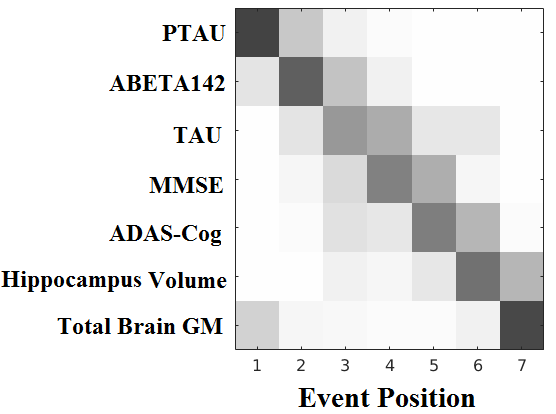}
        \caption{}
    \end{subfigure}
    \caption{ (a) AUC measures in the $10$-folds of cross validation for $7$ events and $47$ events. (b) Positional variance diagram of the central ordering }
    \label{fig:adni}%
\end{figure*}

The plots in Figure~\ref{fig:adni} (a) shows the AUC measures in the $10$-folds of cross validation. The different methods mentioned in the plot indicate the method used for obtaining the event ordering based on which patient staging was done. The results on the left and right side of the graph are for the case of 7 biomarkers and 47 biomarkers respectively. It must be noted that, in the absence of groundtruth event ordering, results using clinical data only provide circumstantial evidence about the reliability of the method and do not unambiguously prove that one method is better than the other.
It can be observed that DEBM based patient staging outperforms both Fonteijn's EBM and Huang's EBM based staging, when used as a classifier. The reduction in AUC while using $47$ biomarkers as compared to $7$ biomarkers indicates that the set of $47$ biomarkers is not optimum for the purpose of classification and an optimum subset selection might be required if this is indeed meant to be used as a classifier. However, as the purpose here is to understand the disease progression mechanism, the decrease in AUC values is not of much significance. The decrease in AUC measure for Huang's EBM is much more than the other two EBMs, when the number of biomarkers increases. This indicates that Huang's EBM might not be scalable.

The plots in Figure~\ref{fig:adni} (b) shows the positional variance diagram of the central ordering while using $100$ sets of bootstrapped samples. Uncertainty in the estimation of central ordering can be observed in this diagram.

\subsection{Simulation Data} \label{ssec:simulres}

\begin{figure}[t!]
\centering
\includegraphics[width=10cm]{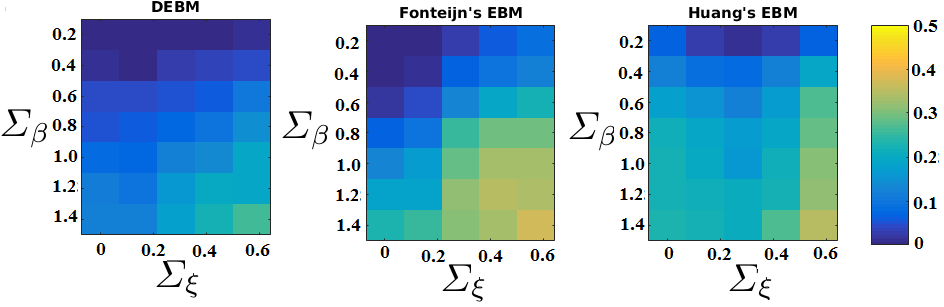}
\caption{Effect of $\Sigma_\beta$ and $\Sigma_\xi$ in the mean inaccuracies of DEBM, Fonteijn's EBM and Huang's EBM. Inaccuracies are measured by the distance of estimated event ordering from the groundtruth ordering. These distances are represented based on the shown colormap.}
\label{fig:VaryNoise7_2D}
\end{figure}

\begin{figure*}[t!]
    \centering
    \begin{subfigure}[b]{1\textwidth}
        \centering
        \label{fig:1}
        \includegraphics[width=1\linewidth]{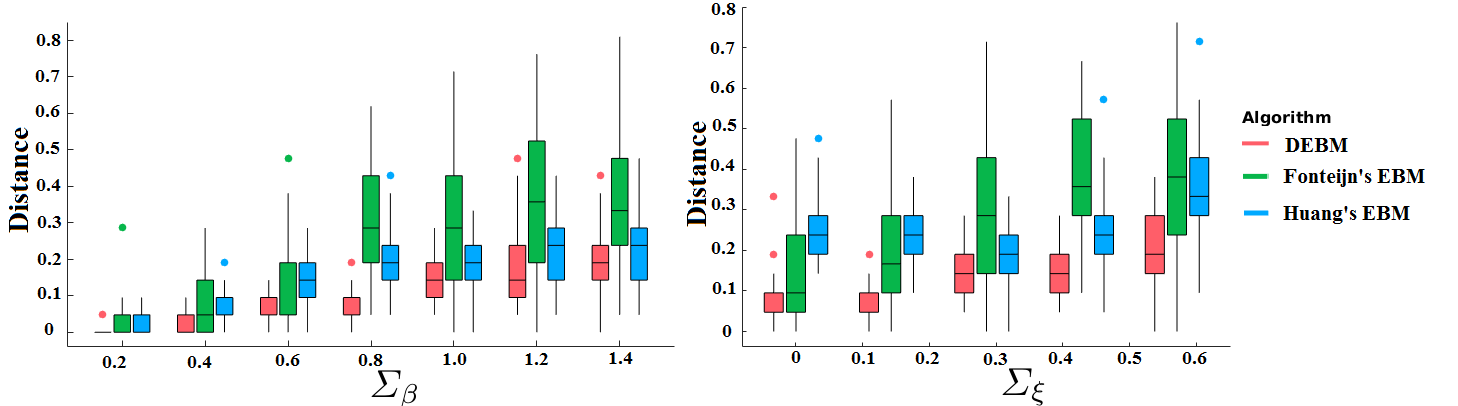}
        \caption{}
    \end{subfigure}%
    ~ \par
    \begin{subfigure}[b]{1\textwidth}
        \centering
        \label{fig:2}
        \includegraphics[width=1\linewidth]{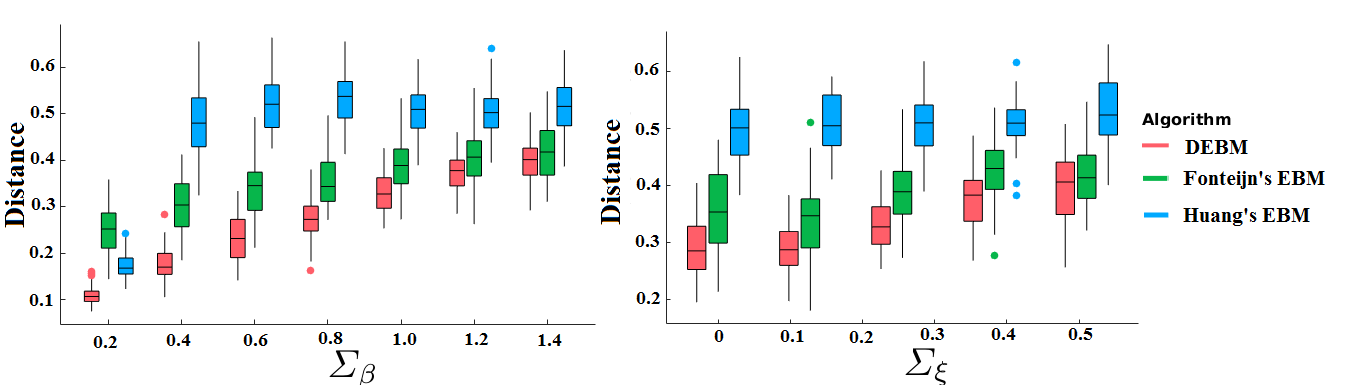}
        \caption{}
    \end{subfigure}
    \caption{Variation in inaccuracies of DEBM, Fonteijn's EBM and Huang's EBM for 50 repetitions of simulations. The number of events is $7$ in (a) and $47$ in (b).}
    \label{fig:VaryNoise_1D}%
\end{figure*}

Figure~\ref{fig:VaryNoise7_2D} shows the variation of mean inaccuracies of DEBM, Fonteijn's EBM and Huang's EBM while varying $\Sigma_\beta$ and $\Sigma_\xi$ of biomarkers. Figures~\ref{fig:VaryNoise_1D} (a) and (b) depicts the variation of inaccuracies for $50$ repetitions of simulated data for the case of $7$ biomarkers and $47$ biomarkers respectively. In Figure~\ref{fig:VaryNoise_1D} (a) for the graph on the left, $\Sigma_\xi$ was fixed to a value of $2/7$, while varying $\Sigma_\beta$. For the graph on the right, $\Sigma_\beta$ was fixed to be $1$ and $\Sigma_\xi$ was varied from $0$ to $4/7$ in steps of $1/7$. Similarly, in Figure~\ref{fig:VaryNoise_1D} (b) for the graph on the left, $\Sigma_\xi$ was fixed to a value of $12/47$, while varying $\Sigma_\beta$. For the graph on the right, $\Sigma_\beta$ was fixed to be $1$ and $\Sigma_\xi$ was varied from $0$ to $24/47$ in steps of $6/47$. Figures~\ref{fig:VaryNoise7_2D}, ~\ref{fig:VaryNoise_1D} (a) and (b) show that DEBM outperforms the state-of-the-art EBM techniques in recovering the order in which biomarkers become abnormal. It can also be seen that, while the performance of DEBM and Fonteijn's EBM was similar for $7$ and $47$ biomarkers, the performance of Huang's EBM degrades with increasing number of biomarkers.

\begin{figure}[t!]
\centering
\includegraphics[width=9cm]{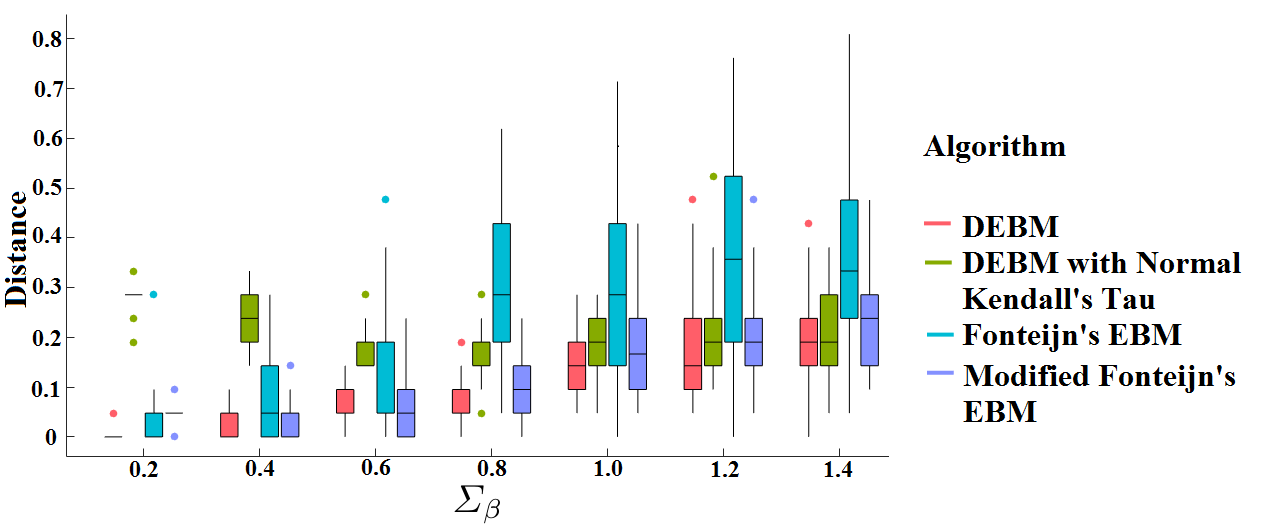}
\caption{Variation in inaccuracies of DEBM, DEBM with Kendall's Tau, Fonteijn's EBM and Modified Fonteijn's EBM with respect to $\Sigma_\beta$, with $\Sigma_\xi = 2/7$.}
\label{fig:OtherEffects}
\end{figure}

Figure~\ref{fig:OtherEffects} shows the results for Experiment $6$ detailed in Section~\ref{ssec:simulexp}. It can be seen that DEBM with probabilistic Kendall's Tau distance outperforms DEBM with normal Kendall's Tau distance. Fonteijn's EBM using normal and abnormal biomarker distributions computed using the technique proposed in Section~\ref{ssec:subjectordering} outperforms conventional Fonteijn's EBM. It is also interesting to note that DEBM outperforms the modified Fonteijn's EBM as well, which uses the same biomarker distribution estimation algorithm as DEBM. 

\begin{figure}[t!]
\centering
\includegraphics[width=6cm]{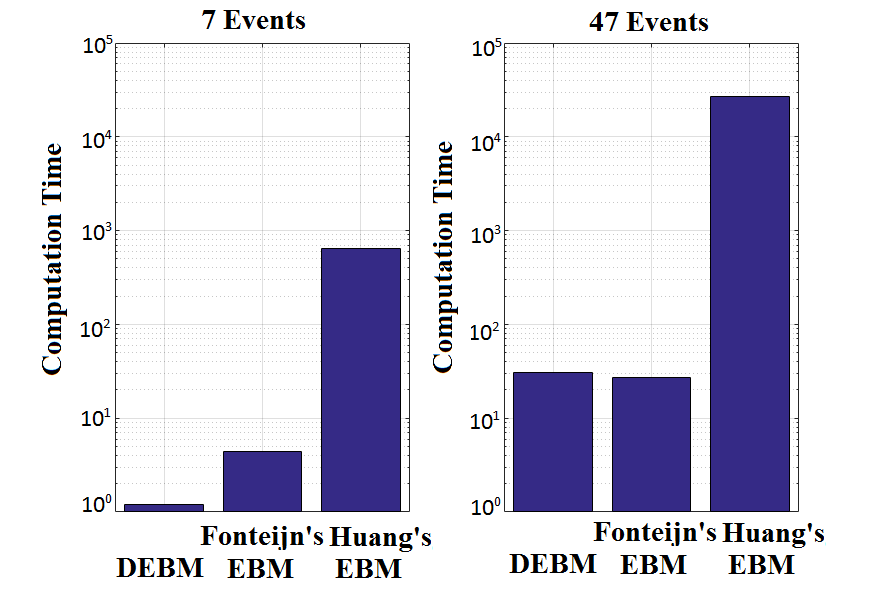}
\caption{Computation Time (in seconds)}
\label{fig:MeanTime}
\end{figure}

Figure~\ref{fig:MeanTime} shows the mean computation time (in seconds) for the three methods in logarithmic scale. The implementation for all the methods were done in Python and measured in the same computer. DEBM is several orders of magnitude faster than Huang's EBM and comparable to Fonteijn's EBM.

\section{Conclusion and Future Work} \label{sec:conc}

We proposed a novel discriminative EBM framework to estimate the ordering in which biomarkers become abnormal during disease progression, based on a cross-sectional dataset. The proposed framework outperforms state-of-the-art EBM techniques in estimating the event ordering and is computationally very efficient as well. In addition to the framework, we also proposed a novel probabilistic Kendall's Tau distance metric and a biomarker distribution estimation algorithm. Each aspect of the proposed algorithm was ascertained to contribute positively in improving the accuracy of estimation.

Fonteijn's EBM assumes a unique event ordering that is common to all the subjects in the database. When this assumption fails, the performance of the algorithm degrades. Huang's EBM and DEBM account for this variation. However, Huang's EBM estimates a lot of parameters through optimization for inferring the central ordering, whereas DEBM is much more direct in estimating the central ordering. This might be one of the reasons why DEBM outperforms Huang's EBM consistently. Moreover, the simplicity of the DEBM algorithm is crucial for its scalability to large number of biomarkers.

Many possible extensions of the work are interesting to consider. Huang's EBM was extended to estimating clusters of central orderings in~\cite{Young:2015} using Dirichlet process mixtures of generalized Mallows models. Such an extension is also possible using DEBM. Subject-specific event orderings estimated based on cross-sectional data are very noisy. Incorporating longitudinal data to get better estimates of subject-specific event orderings in DEBM is also worth considering.

\section*{Acknowledgement}

This work is part of the EuroPOND initiative, which is funded by the European Union's Horizon 2020 research and innovation programme under grant agreement No. 666992. The authors also thank Dr. Jonathan Huang for sharing the implementation of Huang's EBM and Dr. Alexandra Young for the useful discussions on estimation of biomarker distributions as well as for sharing the implementation of the simulation system for biomarker evolution.

\bibliographystyle{splncs03}
\bibliography{EBM}

\end{document}